\documentclass{ifacconf}
\usepackage{placeins}
\usepackage{natbib}
\usepackage{comment}
\usepackage{graphicx}
\usepackage{tabularx,booktabs}
\usepackage{color}
\usepackage{multicol}
\usepackage{amsmath,amssymb}
\usepackage{amsfonts}
\usepackage{textcomp}
\usepackage{psfrag}
\usepackage{multimedia}
\usepackage{fancybox}
\usepackage{amsmath}  % For mathematical expressions

\usepackage{algorithm}
\usepackage{algpseudocode}

\usepackage{amssymb}  % For symbols
\usepackage{bm}       % For bold math symbols
\usepackage{stackengine}
\usepackage{varioref}
\usepackage{import}
\usepackage{framed,xcolor}
\usepackage{color}
\usepackage{subcaption}
\usepackage{float}
\usepackage{etoolbox}
\usepackage{soul}

\definecolor{hos}{rgb}{0.91,0.84,0.42}

\newcommand{\vect}[1]{\ensuremath{\boldsymbol{\mathrm{#1}}}}

\newtheorem{Definition}{Definition}

\begin{document}
\begin{frontmatter}

\title{Safe Robust Predictive Control-based Motion Planning of Automated Surface Vessels in Inland Waterways} 
% Title, preferably not more than 10 words.

\thanks[footnoteinfo]{This work was supported by the US National Science Foundation under award CNS-2302215.}

\author[First]{Sajad Ahmadi} 
\author[First]{Hossein Nejatbakhsh Esfahani} 
\author[First]{Javad Mohammadpour Velni}

\address[First]{Dept. of Mechanical Engineering, Clemson University, Clemson, SC 29634 USA (e-mail: \{sahmadi, hnejatb, javadm\}@clemson.edu).}

\begin{abstract}                % Abstract of not more than 250 words.
Deploying self-navigating surface vessels in inland waterways offers a sustainable alternative to reduce road traffic congestion and emissions. However, navigating confined waterways presents unique challenges, including narrow channels, higher traffic density, and hydrodynamic disturbances. Existing methods for autonomous vessel navigation often lack the robustness or precision required for such environments. This paper presents a new motion planning approach for Automated Surface Vessels (ASVs) using Robust Model Predictive Control (RMPC) combined with Control Barrier Functions (CBFs). By incorporating channel borders and obstacles as safety constraints within the control design framework, the proposed method ensures both collision avoidance and robust navigation on complex waterways. Simulation results demonstrate the efficacy of the proposed method in safely guiding ASVs under realistic conditions, highlighting its improved safety and adaptability compared to the state-of-the-art.
\end{abstract}

\begin{keyword}
Autonomous Surface Vessels, Safety-Critical MPC, Control Barrier Functions, Inland Waterway Transportation.
\end{keyword}

\end{frontmatter}
%===============================================================================
%===============================================================================

\section{Introduction}
\vspace{-2mm}
% Inland water transportation (IWT) offers a sustainable solution to road congestion and the reduction of greenhouse gas emissions. Compared to road systems, waterways significantly lower transportation’s environmental footprint \citep{blasing2016emission}. This potential, coupled with urbanization and logistics demands, has spurred interest in Automated Surface Vessels (ASVs) in IWT. ASVs promise safer and more efficient water transport by eliminating human errors and optimizing routes. However, integrating autonomous navigation technologies into IWT poses unique challenges. Confined waterways have narrow channels, high traffic density, and hydrodynamic phenomena like channel flow effects and boundary interactions. These constraints require specialized sensing and control algorithms for safe and efficient navigation, especially during overtaking or sharp turns.
Inland water transportation (IWT) offers a sustainable solution to road congestion and the reduction of greenhouse gas emissions. Compared to road systems, waterways significantly lower transportation’s environmental footprint \citep{blasing2016emission}. This potential, coupled with urbanization and logistics demands, has spurred interest in Automated Surface Vessels (ASVs) in IWT. ASVs promise safer and more efficient water transport by eliminating human errors and optimizing routes. However, integrating autonomous navigation technologies into IWT poses unique challenges due to confined waterways’ narrow channels, high traffic density, and hydrodynamic phenomena. These constraints require specialized sensing and control algorithms for safe and efficient navigation, especially during overtaking or sharp turns \cite{cheng2021we}.

% Water transportation has emerged as a sustainable solution within the transportation sector, offering an eco-friendly solution to ease traffic congestion and reduce emissions. The key benefit of waterways is their significantly reduced environmental impact compared to road traffic emissions \cite{blasing2016emission}. Since its introduction, there has been a growing demand for water transportation in various sectors, such as cargo shipping, passenger ferries, fishing vessels, cruise ships, and so forth. Consequently, there is a requirement for effective water-based navigation to prevent collisions, especially in congested areas.

A vision-based obstacle avoidance method for ASV navigation was proposed in \cite{kim2019vision}; however, its performance is highly sensitive to environmental conditions, and its ability to detect faraway objects is limited. To improve localization, \cite{dhariwal2007experiments} introduced a multi-sensor fusion approach that enhances the accuracy of the boat’s position estimate. Meanwhile, dedicated efforts have been made in ASV trajectory planning \cite{bitar2020two} and trajectory tracking control \cite{wang2016fast}, yet these methods do not fully tackle the robust safety challenges inherent in dynamic, confined maritime environments.

Model Predictive Control (MPC) has gained widespread adoption for collision avoidance due to its ability to systematically handle both input and state constraints \citep{rawling}. In MPC, safety is typically enforced by incorporating constraints within the optimization problem. However, scenario-based MPC formulations—such as those employing dynamic probabilistic risk assessment \citep{trym2020collision} or incorporating COLREGs \citep{johansen2016ship}—often depend on the accuracy of state estimates and collision risk metrics, leaving them vulnerable under uncertainty. Other approaches, including tube-based MPC with LPV models \citep{yang2023tube} and dynamic programming-based robust control \citep{esfahani2021}, further underscore the challenges of achieving robust, computationally efficient ASV control.

Recent work in nonlinear robust MPC \citep{diehl2010lyapunov, lucia2012new} and sample-based disturbance handling \citep{hewing2019scenario, carvalho2014stochastic} illustrates the potential for advanced optimization techniques to manage uncertainties. Nonetheless, a critical gap remains in explicitly integrating safety constraints into these frameworks for the unique challenges of ASV navigation in confined waterways—a gap that the proposed RMPC-CBF framework seeks to address.

\begin{figure}[H]
    \centering
    \includegraphics[width=.3 \textwidth]{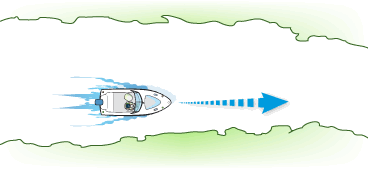}
    \caption{ASVs navigating confined waterways must maintain safe clearance from channel borders.}
    \label{fig:ASV_0}
\end{figure}

To the best of the authors' knowledge, no work on ASV navigation in confined waterways has carefully considered safety issues when obstacle avoidance and safe navigation are combined through narrow channels. As shown in Fig. \ref{fig:ASV_0}, when navigating through a confined channel, ASVs should normally overtake on the port side, while the vessel being overtaken should ideally be positioned as closely as possible to the starboard side of the channel.

\textbf{This paper introduces a new RMPC-CBF framework for safe motion planning of Automated Surface Vessels (ASVs) in confined waterways, addressing the unique challenges posed by narrow channels and high traffic densities. By embedding the channel borders as Control Barrier Functions within the RMPC formulation, our approach effectively maintains safe distances while enabling dynamic maneuvers such as overtaking. This integrated method not only enhances collision avoidance and trajectory tracking under challenging hydrodynamic conditions but also achieves a critical balance between safety and performance, thereby advancing the state-of-the-art in ASV navigation. }Using CBFs has become popular for synthesizing safety-critical controllers due to their generality and relative ease of synthesis and implementation \citep{7782377}. Previous studies have explored the inclusion of safety considerations within the MPC framework to ensure that the controlled system operates within predefined safety boundaries. For example, barrier functions were utilized to develop a safety-critical MPC with CBFs by \citep{zeng2021safety}, \citep{learningCBF2024} and \citep{10566258}.

The paper is organized as follows: Section \ref{sec:2} introduces the dynamic model of the ASV. Section \ref{sec:3} reviews CBFs in continuous and discrete time, setting the stage for our approach. Section \ref{sec:4} details the novel RMPC-CBF framework for safe motion planning in confined waterways. Section \ref{sec:5} presents simulation results demonstrating the framework's effectiveness and performance. Section \ref{sec:6} concludes with outcomes and future research directions.

% A dynamic model of ASVs used in this work is described in Section \ref{sec:2}. In Section \ref{sec:3}, we provide background information on the CBFs in both continuous-time and discrete-time domains. We describe the proposed RMPC-CBF scheme in Section \ref{sec:4}. The simulation results and discussion are provided in Section \ref{sec:5}, which also delves into the simulation of the proposed safety-critical robust model predictive control (RMPC) framework applied to the autonomous surface vehicle's (ASV) motion planning in a confined channel environment. Finally, concluding remarks are made in Section \ref{sec:6}.

\vspace{-2mm}
\section{Modeling of Autonomous Surface Vessels} \label{sec:2}
\begin{figure}
    \centering
    \includegraphics[width=.35 \textwidth]{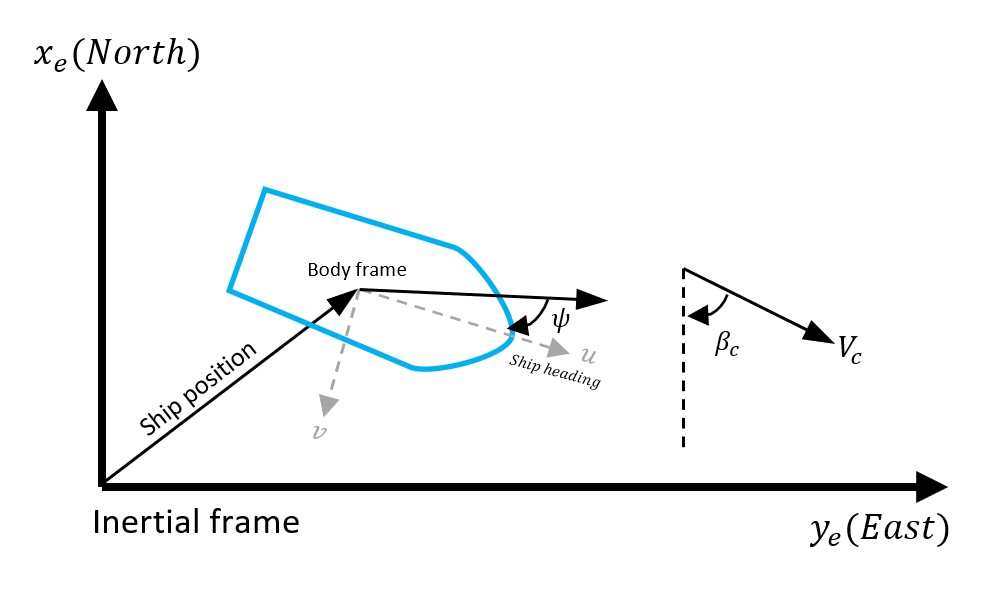}
    \caption{Earth-fixed positioning and body-fixed motion frames for ASV surge, sway, and yaw.}
    \label{fig:ASV_1}
\end{figure}

\begin{figure}
    \centering
    \includegraphics[width=.4 \textwidth]{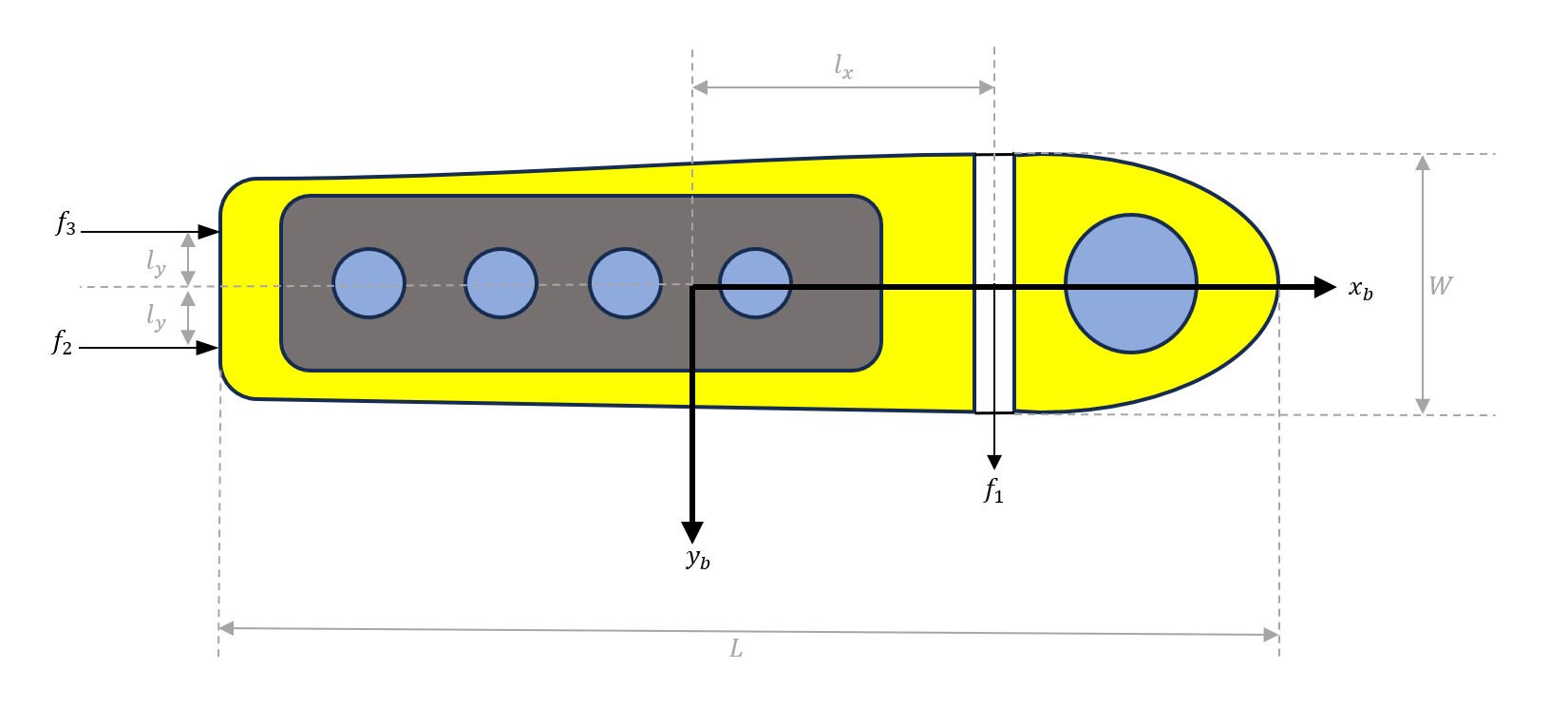}
    \caption{Illustration of the ASV’s rear and side thrusters, showing their placement for generating control forces and torque in confined waterway navigation.}
    \label{fig:ASV_2}
\end{figure}

\vspace{-2mm}
In this paper, we adopt a 3-DOF model of a CyberShip \citep{skjetne2004modeling}. The dynamic is nonlinear and can be described in two reference frames: the body-fixed frame $O_B$ and the Earth-fixed frame $O_E$, as shown in Fig. \ref{fig:ASV_1}. The pose vector $\vect{\eta} = [x,y,\psi]^\top \in \mathbb{R}^3$ is in the Earth-fixed frame, where $x$ is the North position, $y$ is the East position and $\psi$ is the heading angle. The velocity vector $\vect{\nu} = [u,v,r]^\top \in \mathbb{R}^3$ includes the surge $u$ and sway $v$ velocities and yaw rate $r$ in the body-fixed frame. The control inputs and the external time-varying disturbances are labeled as $\vect\tau,\vect\tau_d \in \mathbb{R}^{3}$, respectively. The model can be represented as
\begin{subequations} \label{eq:Dynamic_of_ASV}
    \begin{align}
        &\dot{\vect\eta} = J\left(\vect\eta \right)\vect{\nu},\\
        &M_{RB}\dot{\vect{\nu}}+ M_{A}\dot{\vect{\nu}} +C_{RB}\left(\vect{\nu}\right)\vect{\nu}\\\nonumber
        &\qquad\qquad+C_{A}\left(\vect{\nu}\right)\vect{\nu} + D\left(\vect{\nu}\right)\vect{\nu} =\vect\tau + \vect \tau_d,
    \end{align}
\end{subequations}
where the rotation matrix $J(\eta)$ is defined as
\begin{align}
    J\left(\vect\eta \right) = \begin{bmatrix}
    \cos\left(\psi\right) & -\sin\left(\psi \right) & 0\\
    \sin\left(\psi \right) & \cos\left(\psi \right) & 0\\
    0 & 0 & 1\end{bmatrix}.
\end{align}
The rigid-body inertia matrix $M_{RB}$ and the added mass $M_A$ are computed as 
\begin{align}
    M_{RB} = \begin{bmatrix}
    m & 0 & 0\\
    0 & m & mx_g\\
    0 & mx_g & I_z\end{bmatrix},\quad M_{A} = \begin{bmatrix}
    -X_{\dot{u}} & 0 & 0\\
    0 & -Y_{\dot{v}} & -Y_{\dot{r}}\\
    0 & -N_{\dot{v}} & -N_{\dot{r}}\end{bmatrix},
\end{align}
where $m$ is the mass of the ASV, $I_z$ is the moment of inertia about the body $z_b$-axis (yaw axis), and $x_g$ is the distance between the center of gravity and the body $x_b$-axis. Also, the rigid body and hydrodynamic of the Centripetal and Coriolis acceleration matrices are specified as
\begin{align}
    C_{RB}\left(\vect\nu \right) = \begin{bmatrix}
    0 & 0 & -m(x_gr+v)\\
    0 & 0 & mu\\
    m(x_gr+v) & -mu & 0\end{bmatrix},
\end{align}

\begin{align}
    C_{A}\left(\vect\nu \right) = \begin{bmatrix}
    0 & 0 & c_{13}\\
    0 & 0 & c_{23}\\
    -c_{13} & -c_{23} & 0\end{bmatrix},
\end{align}
with $c_{13} = Y_{\dot{v}}v_r + 0.5(N_{\dot{v}} + Y_{\dot{r}})r, ~c_{23} = -X_{\dot{u}}u_r $, and $X_{\dot{u}}, Y_{\dot{v}}, Y_{\dot{r}}, N_{\dot{v}} $ and $N_{\dot{r}}$ are constant model parameters, \textcolor{black}{while $u_r$ and $v_r$ are the reference frame surge and sway velocities, respectively.} Moreover, the damping matrix reads as
\begin{align}
    D\left(\vect\nu \right) = \begin{bmatrix}
    d_{11}& 0 & 0\\
    0 & d_{22} & d_{23}\\
    0 & d_{32} & d_{33}\end{bmatrix},
\end{align}
where
\begin{subequations} \label{}
    \begin{align}
        &d_{11} = X_u + X_{|u|u}|u| + X_{uuu}{u_r}^2,\\
        &d_{22} = Y_v + Y_{|v|v}|v_r| + Y_{|r|v}|r|,\\
        &d_{23} = Y_r + Y_{|v|r}|v_r| + Y_{|r|r}|r|,\\
        &d_{32} = N_v + N_{|v|v}|v_r| + N_{|r|v}|r|,\\
        &d_{33} = N_r + N_{|v|r}|v_r| + N_{|r|r}|r|,
    \end{align}
\end{subequations}
where the hydrodynamic coefficients are specified by $X_{(.)}, Y_{(.)}$ and $N_{(.)}$. Also, $\vect{\tau} = [X,Y,N]^{\top}$ includes the external control forces $X, Y$ and the moment vector $N$. The thrusters are located in the back and the side of the ship, as depicted in Fig. \ref{fig:ASV_2}.
\textcolor{black}{%
The actuator forces/torque vector is $\vect{F} = [f_1,f_2,f_3]^\top$, and the control input and the external disturbance vector are obtained as
\begin{subequations}
    \begin{align}
        \vect{\tau} &= \mathbf{B}_T \vect{F},\\[1mm]
        \vect{\tau}_d &= \mathbf{B}_T \vect{\omega}_d,
    \end{align}
\end{subequations}
where $\vect{\omega}_d \in \mathbb{R}^3$ is the disturbance vector. The thruster allocation matrix is given by
\begin{align}
    \mathbf{B}_T = \begin{bmatrix}
    0 & 0 & 1\\[1mm]
    1 & 0 & 0\\[1mm]
    l_x & -l_y & l_y
    \end{bmatrix},
\end{align}
with $l_x$ and $l_y$ denoting the offsets from the vessel’s center of gravity to the thrusters in the longitudinal and lateral directions, respectively (see Fig.~\ref{fig:ASV_2}). This configuration ensures that the spatial arrangement of the thrusters is accurately incorporated into the resulting control input.
}

% The model and the corresponding parameters used in this work are borrowed from \citep{skjetne2004modeling}. %Nevertheless, we plan to investigate the integration of robust MPC schemes with CBFs in our future work to address this issue. 
 
% \textcolor{blue}{Here, $l_x$ and $l_y$ denote the offsets from the vessel's center of gravity to the thrusters in the longitudinal and lateral directions, respectively (see Fig.~\ref{fig:ASV_2}). This configuration guarantees that the spatial arrangement of the thrusters is faithfully incorporated into the resulting control input.}

% \textcolor{blue}{In addition, the effect of external disturbances is incorporated into the system dynamics via a disturbance torque term:

% \begin{align} \label{eq:dist}
%     \vect{\tau}_d = \begin{bmatrix}
%     0 & 0 & 1\\
%     1 & 0 & 0\\
%     l_x & -l_y & l_y\end{bmatrix}\vect \omega_d.
% \end{align}

% where $\vect{\omega}_d \in \mathbb{R}^3$ is the disturbance vector.
% }

\section{REVIEW OF CONTROL BARRIER FUNCTIONS} \label{sec:3}
\vspace{-2mm}
Control barrier functions (CBFs) have been proposed to ensure safety in safety-critical systems by restricting the control inputs to an acceptable range. This is achieved by defining safety requirements as the need for a system to remain within a defined safe set. To accomplish this, we define a set $\mathcal{C}$ as the region above a continuously differentiable function $h:\mathcal{D}\subset\mathbb{R}^n\rightarrow\mathbb{R}$, where $\mathcal{D}$ represents a subset of $n$-dimensional real numbers such that
\begin{align}\label{eq:safe_set}
    &\mathcal{C}=\left\{\vect x\in\mathcal{D}\subset\mathbb{R}^n: h\left(\vect x\right)\geq 0\right\},\\\nonumber
    &\partial\mathcal{C}=\left\{\vect x\in\mathcal{D}\subset\mathbb{R}^n: h\left(\vect x\right)=0\right\},\\\nonumber
    &\text{Int}\left(\mathcal{C}\right)=\left\{\vect x\in\mathcal{D}\subset\mathbb{R}^n: h\left(\vect x\right)> 0\right\},
\end{align}
where $\partial\mathcal{C}$ and $\text{Int}\left(\mathcal{C}\right)$ are the boundary of $\mathcal{C}$ and the interior of $\mathcal{C}$, respectively. In addition, we make the assumption that the interior of the set $\mathcal{C}$ is not empty (denoted as $\text{Int}(\mathcal{C})\neq\emptyset$). We then define $\mathcal{C}$ as the ``safe set." A CBF is considered successful in certifying forward invariance of $\mathcal{C}$ if it ensures that the system trajectory remains inside the safe set and does not approach its boundary. Let us consider a nonlinear system in control-affine form as
\begin{align}\label{eq:sys}
    \dot{\vect x}=\vect f\left(\vect x\right)+ g\left(\vect x\right)\vect u,
\end{align}
where $\vect f: \mathbb{R}^n\rightarrow\mathbb{R}^n$ and $g: \mathbb{R}^n\rightarrow\mathbb{R}^{n\times m}$ are locally Lipschitz continuous functions, $\vect x\in\mathbb{R}^n$ and $\vect u\in\mathbb{R}^m$ are the system states and control inputs, respectively. The closed-loop dynamic of the system is then described as
\begin{align}\label{eq:closed_loop}
    \dot{\vect x}=\vect f_{\text{cl}}\left(\vect x\right)=\vect f\left(\vect x\right)+g\left(\vect x\right)\vect u|_{\vect\pi\left(\vect x\right)},
\end{align}
where the control policy (feedback controller) $\vect\pi:\mathbb{R}^n\rightarrow\mathbb{R}^m$ is locally Lipschitz continuous. Consider a maximum interval of existence denoted as $I(\mathbf{x_0}) = [t_0, t_{\text{max}})$ for any initial condition $\mathbf{x_0}$ in the subset $\mathcal{D}$. This interval ensures that the trajectory $\mathbf{x}(t)$ represents the unique solution to \eqref{eq:closed_loop} within the interval $I(\mathbf{x_0})$. If $t_{\text{max}}$ is set to infinity, the closed-loop system $\mathbf{f_{\text{cl}}}$ is considered forward complete, implying that the system's solutions can be extended indefinitely into the future.

\begin{Definition}{\citep{ames2019control}}~(Forward Invariance)
The closed-loop system \eqref{eq:closed_loop} is forward invariant w.r.t the set $\mathcal{C}$ if for every $\vect x_0\in\mathcal{C}$, we have $\vect x\left(t\right)\in\mathcal{C}$ for all $t\in I\left(\vect x_0\right)$.
\end{Definition}

\begin{Definition}{\citep{ames2019control}}~(Control Barrier Function)
Given a dynamical system \eqref{eq:sys} and the safe set $\mathcal{C}$ with a continuously differentiable function $h:\mathcal{D}\rightarrow\mathbb{R}$, then $h$ is a CBF if there exists a class $\mathcal{K}_\infty$ function $\kappa$ for all $\vect x\in\mathcal{D}$ such that 
     \begin{equation}\label{eq:cbf0}
         \sup_{\vect u\in\mathcal{U}}\left\{L_fh\left(\vect x\right)+L_gh\left(\vect x\right)\vect u\right\}\geq-\kappa\left(h\left(\vect x\right)\right),
     \end{equation}
where $\dot h\left(\vect x,\vect u\right)=L_fh\left(\vect x\right)+L_gh\left(\vect x\right)\vect u$ with $L_fh,L_gh$ as the Lie derivatives of $h$ along the vector fields $\vect f$ and $g$. A common choice for the function $\kappa$ is a linear form given by $\kappa(h(\mathbf{x})) = \alpha h(\mathbf{x})$, where the parameter $\alpha \geq 0$ influences the system behavior near the boundary of $h(\mathbf{x}) = 0$.
\end{Definition}

The above condition in discrete time becomes
 \begin{align}\label{eq:discrete_CBF}
    \Delta h\left(\vect x_k\right) \geq -\gamma h \left(\vect x_k\right), ~~~0<\gamma\leq 1,
 \end{align}
 where $\Delta h\left(\vect x_k\right) :=h\left(\vect x_{k+1}\right)-h\left(\vect x_k\right)$.

\section{Safety-Critical Control Design} \label{sec:4}
\vspace{-2mm}
In this section, we first formulate the safety functions required in the discrete-time CBFs \eqref{eq:discrete_CBF} to address a safe motion planning of ASVs in confined inland waterways. We then formulate the safety-critical RMPC scheme based on the proposed CBFs. 

\subsection{Safety Functions}
This paper addresses two types of unsafe zones formulated as discrete-time CBFs. The first type describes some obstacles $i=1,\ldots,n_o$ for which the safety function  $h_i^o\left(x_k,y_k\right)$ is defined as  
\begin{align}
h_i^o\left(x_k,y_k\right):=-1 +\frac{\left( {\left(x_k-o_{x,i}\right)^2+\left(y_k-o_{y,i}\right)^2} \right)}{{{\left( {{r_i} + {r_a}} \right)}^2}}\geq 0,
\end{align}
where $(o_{x,i},o_{y,i})$ and $r_i$ are the center and radius of the $i^{\mathrm{th}}$ obstacle, respectively. The constant $r_a$ is the radius of the ASV. The second type of unsafe zone is considered for the borders of the inland waterway. To this end, we define these borders by two straight lines $L_{j}$, where $j=1,2$. Each line is then characterized by 
\begin{align}
L_{j}:=a_{j}x_k+b_{j}y_k+c_{j}=0,\quad j=1,\ldots,z
\end{align}
where $z$ is the number of border lines and $a_{\ell,j},b_{\ell,j},c_{\ell,j}\in\mathbb{R}$ are the parameters of each line. The corresponding safe set is then obtained as
\begin{align}
    \mathcal{C}_b=\left\{\vect P_k\in\mathbb{R}^2|L_1\geq0\cap L_2\geq0\right\},
\end{align}
where $\vect P_k=[x_k,y_k]^\top$ is the 2D position vector of the ASV. Let us define the distance between the ASV and each waterway border as
\begin{align}\label{}
    D_{j}\left(\vect P_k\right)= \frac{a_{j}x_k+b_{j}y_k+c_{j}}{\sqrt{a^2_{j}+b^2_{j}}}.
 \end{align}
The safety functions $h_j^b\left(\vect P_k\right)$ used in the corresponding CBFs are then defined by
\begin{align}\label{}
    &h_j^b\left(\vect P_k\right):= D_{j}\left(\vect P_k\right) - \frac{\sqrt{w^2+l^2}}{2},
 \end{align} 
where $w$ and $l$ denote the width and length of the ASV, respectively.

\subsection{Design of Robust Safety-Critical RMPC Scheme}
Let $\dot{\vect x}=\bar{\vect f}\left(\vect x,\vect u, \vect \omega^r\right)$ be the continuous-time, state-space description of the ASV, where $\vect x=\left[\vect\eta,\vect\nu\right]^\top$ and $\vect \omega^r \in\mathbb{R}^{3}$ is the disturbance applied to the real system. The discretized model then reads a $\vect x_{k+1}=\vect f(\vect x_k,\vect u_k, \vect \omega_k^r)$. We assume that a full measurement or estimate of the state vector $\vect {x_k}$ is available at the current time instant $k$. We then design our safety-critical controller, addressing the motion planning of ASVs while ensuring safety issues in inland waterways. The proposed safety-critical RMPC solves the following constrained finite-time Optimal Control Problem (OCP) with a prediction horizon of length $N$ at each time instant $k$.
\noindent\makebox[\linewidth]{\rule{\linewidth}{0.4pt}}
\noindent {\bf Proposed Robust MPC-CBF Problem}:
\begin{subequations} \label{eq:MPC_CBF}
	\begin{align}
	   \label{eq:cost}
       &J = \min_{\vect x,\vect u} \left\{V\left(\vect x_N\right) + \max_{\vect\omega} \left\{\sum_{k=0}^{N-1} L\left(\vect x_k,\vect u_k, \vect \omega_k\right)\right\}\right\}\\ \label{eq:equality_const}
	   &\mathrm{s.t.}\\\nonumber
	   &{\vect x}_{k+1}=\vect f \left({\vect x}_{k}, {\vect u}_{k}, {\vect \omega}_{k}\right),\\ 
     &\vect g\left({\vect u}_k\right)\leq 0,\\
     &\vect q\left(\vect x_k,{\vect u}_k\right)\leq 0,\quad \vect q_f\left(\vect x_N,{\vect u}_N\right)\leq 0,\\
     &\vect \omega_{min} \leq \vect \omega_k \leq \vect \omega_{max},\\
     &\Delta h_i^o\left(\vect x_k\right)+\gamma_i^o h_i^o\left(\vect x_k\right)\geq 0,\quad i=1,\ldots,n_o,\label{eq:ho}\\
     &\Delta h_j^b\left(\vect x_k\right)+\gamma_j^b h_j^b\left(\vect x_k\right)\geq 0,\quad j=1,2. \label{eq:hb}
	\end{align}
\end{subequations}
\noindent\makebox[\linewidth]{\rule{\linewidth}{0.4pt}}
In \eqref{eq:cost}, $V$ and $L$ denote the terminal and stage costs, respectively. The control input constraints are introduced by $\vect g\left({\vect u}_k\right)$ while $\vect q_f$ and $\vect q$ are the mixed terminal and stage inequality constraints, respectively. The first element of the optimal control input sequence $\vect{u}_{0,\ldots,N-1}^\star$ is then applied to the ASV, and the above OCP is solved at each time instant based on the latest state of the system. The disturbance vector $\vect\omega_k = \left[\omega_x, \omega_y, 0\right]^\top$ is bounded as $ \underline{\omega}_x\leq\omega_{x}\leq \bar{\omega}_x$ and $\underline{\omega}_y\leq\omega_{y}\leq\bar{\omega}_y$, where $\omega_x$ and $\omega_y$ denote the disturbance in $x$ and $y$ directions, respectively. The constraints \eqref{eq:ho} and \eqref{eq:hb} denote the proposed CBF conditions on the obstacles $1, \ldots , n_o$ and borders $1,2$, respectively, \textcolor{black}{and are enforced at every time step within the prediction horizon.} In this paper, we consider a quadratic form for $V$ and $L$ as described in the next section. The proposed approach aims to solve the robust model predictive control (RMPC) problem by minimizing the worst-case cost under bounded disturbances. Let the disturbance inputs $\omega_x$ and $\omega_y$ be quantized into $N_\omega$ levels. At each time instant $k$, we select the disturbance value $\omega_k$ that maximizes the stage cost function
\begin{equation}
    \vect\omega_k = \arg\max_{\omega \in \mathcal{W}} L\left(\vect x_k,\vect u_k, \vect \omega_k\right),
\end{equation}
where $\mathcal{W}$ represents the set of all possible discrete disturbance values, defined as  
\begin{equation*}
    \mathcal{W}_x = \{ \omega_x^1, \omega_x^2, \dots, \omega_x^{N_\omega} \}, \quad
    \mathcal{W}_y = \{ \omega_y^1, \omega_y^2, \dots, \omega_y^{N_\omega} \},
\end{equation*} 
such that each $\omega_x^{(.)}$ and $\omega_y^{(.)}$ is a quantized level within the bounded range $[\underline{\omega}_x, \bar{\omega}_x]$ and $[\underline{\omega}_y, \bar{\omega}_y]$, respectively. The overall disturbance set is then $\mathcal{W} = \mathcal{W}_x \times \mathcal{W}_y$. \textcolor{black}{Then, we incorporate a conditional constraint for each quantized disturbance level, which forces the optimizer to select the disturbance level that results in the maximum stage cost $L$ over $\omega$. Finally, the cost $J$ will be minimized under the constraints.} Algorithm \ref{alg:min_max} outlines the proposed approach in a more organized way. 

% an approach for solving the robust model predictive control (RMPC) problem, where the primary objective is to minimize the worst-case (maximum) cost due to bounded disturbances. Since it is challenging to directly nest the max operation within the min operation in an optimization framework, we circumvent the issue by discretizing the disturbance range into a finite set of values. Specifically, the disturbance interval is divided into $N_\omega$ segments using two linspace vectors, $w_1$ and $w_2$, to represent possible disturbance values. Within the prediction horizon, at each time step $k$, if $k$ equals the final time step $N$, the cost $J_k$ is set to the terminal cost $\min V(\vect x_k)$, effectively concluding the optimization process. Otherwise, an iterative process determines the disturbance that maximizes the cost. The stage cost $L(\vect x_k, \vect u_k, \vect w_k)$ is calculated for each disturbance combination. The maximum stage cost, $\text{cost}_{\max}$, is identified along with the corresponding worst-case disturbance $w_{\text{worst}}$. This maximum disturbance-based cost is then used to compute the smallest value of $J_k$ over $w_{\text{worst}}$, ensuring that the algorithm accounts for the most adverse effect of disturbances in a discrete, computationally manageable way.

% \subsection{Proposed Algorithm and Analysis}

% Algorithm \ref{alg:min_max} shows 

% the prediction horizon in the Robust min-max MPC.

\begin{algorithm}
    \caption{Implementation of Robust MPC Scheme}
    \label{alg:min_max}
    \begin{algorithmic}[1]
        \State Get initial position and final destination of the ASV ($\vect{x}_0$ and $\vect{x}_f$)
        \State $J \gets 0$
        \State $\text{tol} \gets 0.3$
        \While{$\| \vect{x} - \vect{x}_f\|_2 \geq \text{tol}$}
            \For{$k = 0 \to N-1$}
                \State $\vect{\omega}^k_{\text{worst}} \gets [0,0]^\top$
                \State $L^k_{\max} \gets 0$
                \For{$\omega_x \in \mathcal{W}_x$}
                    \For{$\omega_y \in \mathcal{W}_y$}
                        \State $L^k(\vect{x}_k, \vect{u}_k) \gets L(\vect{x}_k, \vect{u}_k, [\omega_x, \omega_y, 0])$
                        \If{$L^k(\vect{x}_k, \vect{u}_k) > L^k_{\max}$}
                            \State $L^k_{\max} \gets L^k(\vect{x}_k, \vect{u}_k)$
                            \State $\vect{\omega}^k_{\text{worst}} \gets [\omega_x,\omega_y,0]^\top$
                        \EndIf
                    \EndFor
                \EndFor
                \State $J \gets J + L^k_{\max}$
            \EndFor
            \State $J \gets J + V(\vect{x}_N)$
            \State Solve the problem of minimizing cost $J$ subject to \eqref{eq:equality_const}-\eqref{eq:hb} to find $\vect{u}$
            \eqref{eq:MPC_CBF}.
            \State Apply $\vect{u}_{0}^\star$ to real system $\vect x_{k+1}=\vect f(\vect x_k,\vect u_k, \vect \omega_k^r)$.
        \EndWhile
    \end{algorithmic}
\end{algorithm}

\textbf{Remark 1:} \textit{We note that the main contribution of this work is in the application of RMPC-CBF algorithm in ASVs. A parallel work is exploring theoretical guarantees for the stability and recursive feasibility of the proposed predictive control scheme.}

\section{Simulation Results and Discussion}\label{sec:5}
\vspace{-2mm}
In this section, we examine the performance of the proposed safety-critical RMPC for  ASV's motion planning. In this scenario, an ASV must safely navigate among static obstacles while avoiding the boundaries of the inland waterway and approaching the final destination. To examine the robustness of the proposed safety-critical RMPC, we consider some external disturbances induced by water flow. The water flow force in the channel is represented by the vector $\vect d_f = \{f, \beta\}$, where $f$ is the flow force and $\beta$ is the flow angle. In our simulations, the flow force $f$ is generated as
\begin{align}
  f = \sqrt{\sin^2(x + y) + \sin^2(x - y)},  
\end{align}

and the flow angle $\beta$ is defined as
\begin{align}
    \beta = \tan^{-1} \left( \frac{\sin(x - y)}{\sin(x + y)} \right),
\end{align}

where $x$ and $y$ are the first two states of the system, respectively. The disturbances acting on the ASV are $\omega_x =  f\cos(\beta)$ and $\omega_y = f\sin(\beta)$, which will be added to the control inputs $f_{surge}$ and $f_{sway}$, respectively.

We select a sampling time of $0.2$ sec. to discretize the continuous-time ASV model and set the prediction horizon to $N=10$. The terminal and stage costs in the RMPC-CBF scheme \eqref{eq:MPC_CBF} are defined as
\begin{align}
    &V\left(\vect x_N\right)=\left(\vect x_N-\vect r_d\right)^\top Q_T \left(\vect x_N-\vect r_d\right),\\
    &L\left(\vect x_k,\vect u_k, \vect \omega_k \right)=\\\nonumber
    &\left(\vect{x}_k-\vect r_d\right)^\top Q \left(\vect{x}_k-\vect r_d\right)+ \left(\Delta \vect{\tilde{u}}_k\right)^\top R\left(\Delta \vect{\tilde{u}}_k\right)
\end{align}
where $\vect r_d$ is the desired state vector, $\Delta \vect{\tilde{u}}_k = \vect{\tilde{u}}_k - \vect{\tilde{u}}_{k-1}$, and $\vect{\tilde{u}}_k = \vect u_k -  [\omega_{x,k};\omega_{y,k};0]$ . The corresponding weights are then selected as $Q_T=\text{diag}\left([3,3,3,1,1,1]\right),Q=\text{diag}\left([2,2,2,1,1,1]\right), R=\text{diag}\left([0.1,0.1,0.01]\right)$,
and other parameters are $N_\omega = 20, w_{min} = -\sqrt{2} , w_{max} = \sqrt{2}, u_{min} = [-8, -8, -8], u_{max} = [8, 8, 8]$.

To solve the RMPC optimization problem, we employed CasADi with IPOPT as the solver \citep{andersson2019casadi}. \textcolor{black}{The motion planning scenario aims to safely guide the ASV from the initial position $P_i = [0,2,0]$, where it starts at rest, to the final destination $P_f = [25, 3, 0]$, where it comes to a complete stop.}

\begin{figure}[htbp!]
\centering
\includegraphics[width=0.3\textwidth]{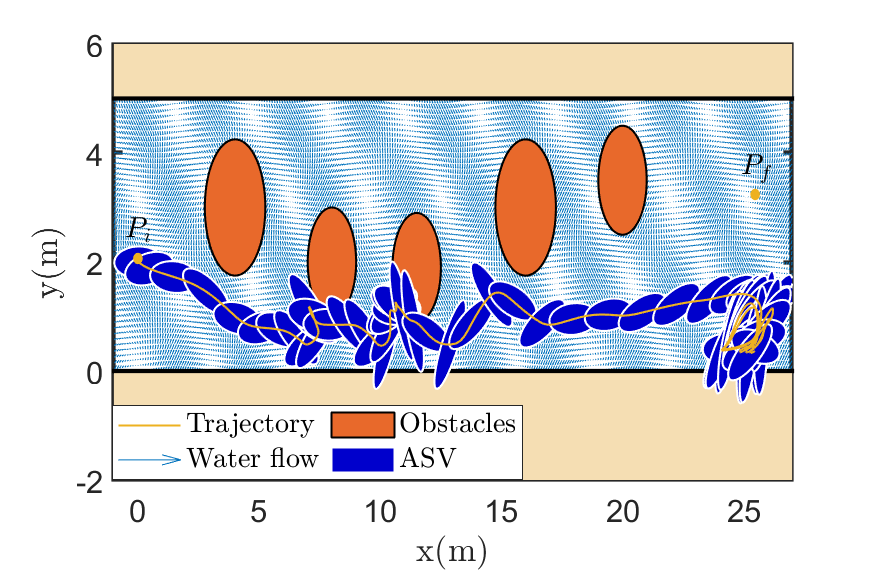}
\caption{The figure shows the ASV's trajectory with a non-robust MPC. Starting at $P_i$, the ASV struggles to navigate through the region, which includes water flows (blue arrows) and obstacles (orange shapes). ASV trajectory under non-robust MPC control shows collisions with obstacles and borders due to the inability to handle disturbances and maintain constraints.} 
\label{fig:MPC}
\end{figure}

\begin{figure}[h!]
    \centering
    % First subfigure
    \begin{subfigure}[t]{0.3\textwidth} % Adjust width as needed
        \centering
        \includegraphics[width=\textwidth]{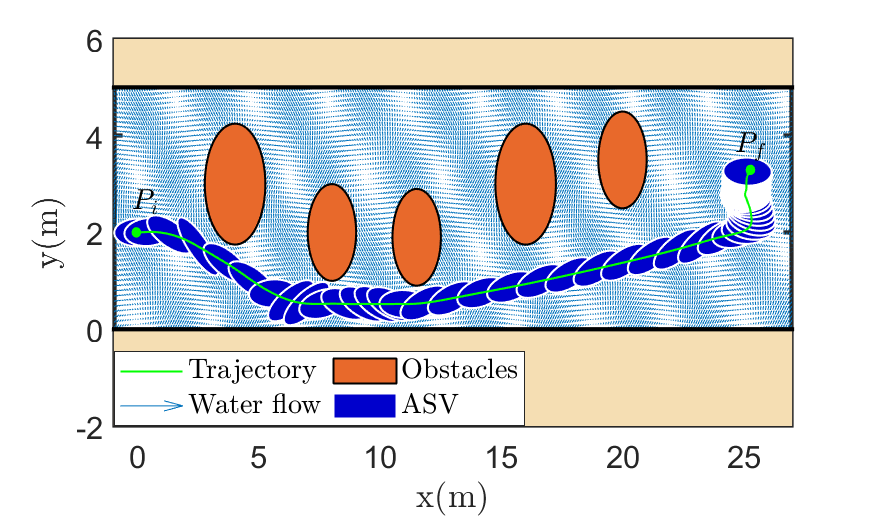} % Replace with your file
        \caption{}
        \label{fig:RMPC}
    \end{subfigure}
    % Add horizontal space
    % \hspace{0.05\linewidth}
    % Second subfigure
    \begin{subfigure}[t]{0.3\textwidth} % Adjust width as needed
        \centering
        \includegraphics[width=\textwidth]{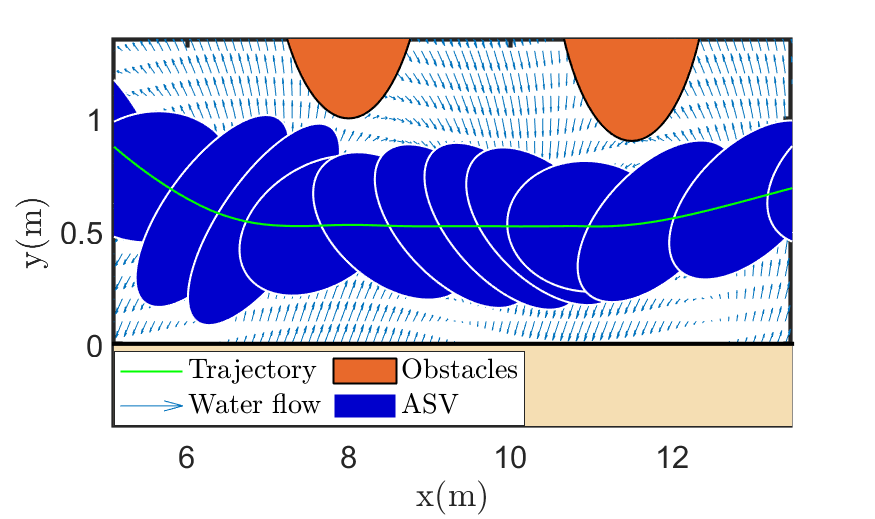} % Replace with your file
        \caption{}
        \label{fig:zoomed}
    \end{subfigure}
    \caption{(a)The green trajectory starts from $P_i$, showing safe navigation using the proposed safety critical RMPC when we have the CBF on the upper and lower borders and all the obstacles with $\gamma^o = 0.15$ and $\gamma^b=0.9$. The safe trajectory of the ASV under the proposed RMPC-CBF framework demonstrates successful obstacle and border avoidance in a confined waterway influenced by disturbances. (b) A zoomed view of the ASV's trajectory shows precise obstacle avoidance and safe navigation along the waterway borders using the proposed RMPC-CBF control.}
\end{figure}

Fig. \ref{fig:MPC} shows the ASV's trajectory when employing a conventional (non-robust) MPC scheme. Due to the influence of water flow disturbances (shown by blue arrows) and the presence of static obstacles (shown as orange shapes), the ASV struggles to follow a safe path. With the lack of robustness in disturbance handling, the ASV collides with both the obstacles and the channel boundaries, making this method unsuitable for safe navigation.  In contrast, Fig. \ref{fig:RMPC} demonstrates the effectiveness of the proposed RMPC-CBF framework. The ASV's trajectory, represented in green, successfully avoids obstacles while remaining within the channel boundaries. The incorporation of the CBFs ensures a collision-free navigation by imposing safety constraints on both obstacles and waterway borders. Additionally, Fig. \ref{fig:zoomed} shows a close-up view of the ASV's trajectory, highlighting its ability to maneuver through the waterway without violating safety constraints.  

\begin{figure}[htbp!]
\centering
\includegraphics[width=0.3\textwidth]{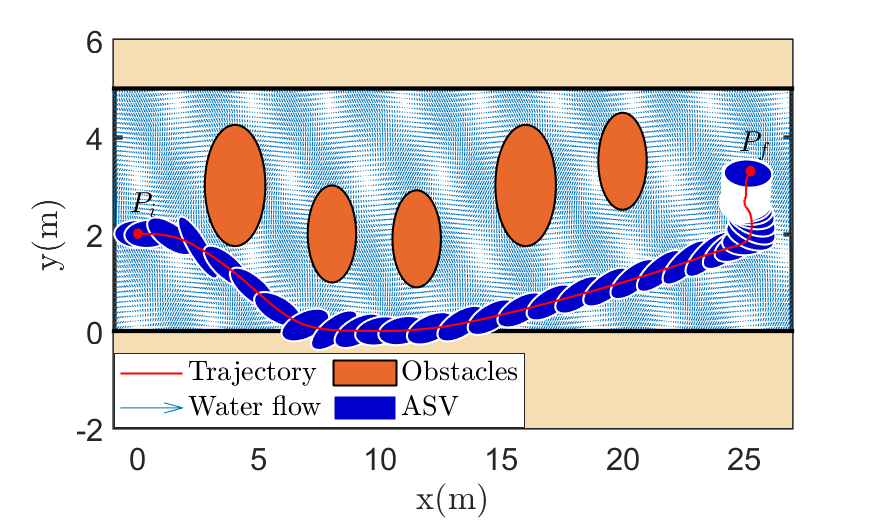}
\caption{\textcolor{black}{ASV trajectory under robust MPC employing hard state constraints for border avoidance. While these constraints effectively prevent the vessel from exiting the channel, they do not offer the adaptive safety margins of the CBF-based approach, resulting in inadequate clearance from the borders during sharp turns and occasional contact with the channel boundaries.}} 
\label{fig:RMPC_No_CBF}
\end{figure}

\textcolor{black}{Fig. \ref{fig:RMPC_No_CBF} shows that while enforcing hard state constraints ensures that the ASV remains within the channel boundaries, this approach does not adjust for varying operational conditions, resulting in insufficient clearance from the channel edges during sharp maneuvers.
}

\begin{figure}[h!]
    \centering
    % First subfigure
    \begin{subfigure}[t]{0.46\linewidth} % Adjust width as needed
        \centering
        \includegraphics[width=\linewidth]{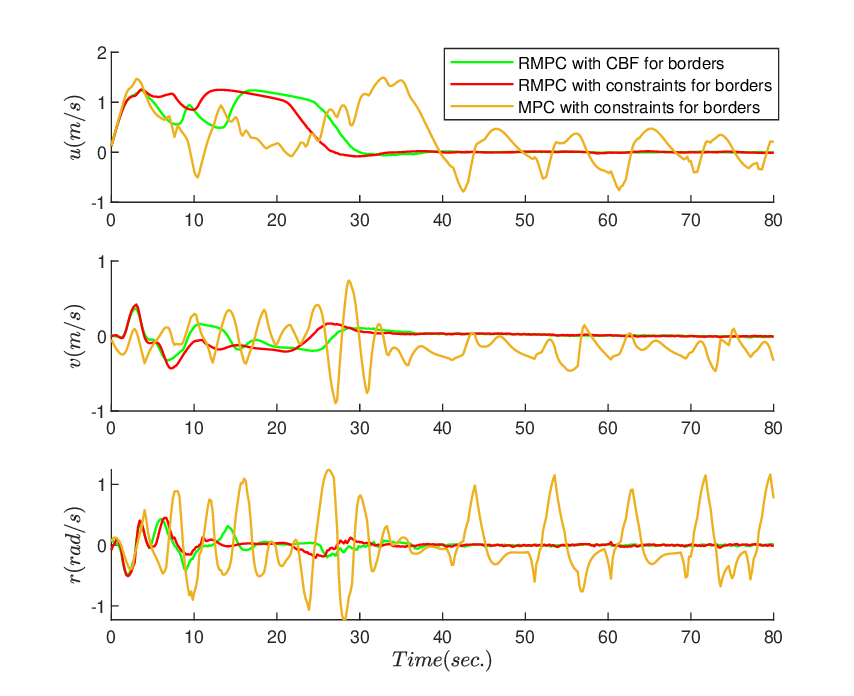} % Replace with your file
        \caption{}
        \label{fig:speed_compare}
    \end{subfigure}
    % Add horizontal space
    % \hspace{0.05\linewidth}
    % Second subfigure
    \begin{subfigure}[t]{0.49\linewidth} % Adjust width as needed
        \centering
        \includegraphics[width=\linewidth]{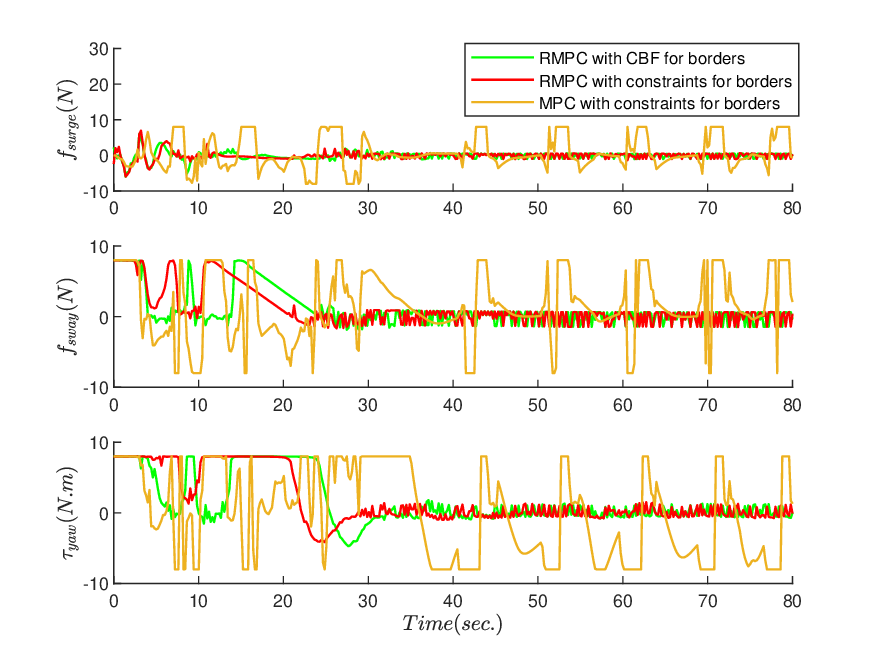} % Replace with your file
        \caption{}
        \label{fig:force_compare}
    \end{subfigure}
    \caption{(a) Comparison of ASV speeds under RMPC-CBF, RMPC with border constraints, and non-robust MPC. (b) Control inputs comparison for ASV navigation using RMPC-CBF, RMPC with border constraints, and non-robust MPC, demonstrating the smoother and more stable control achieved with RMPC.}
    \label{fig:3D_view}
\end{figure}

Finally, Fig. \ref{fig:speed_compare} and \ref{fig:force_compare} compare the speed and control inputs across different scenarios. The results indicate that the RMPC-CBF approach ensures smoother and safer transitions in velocity and control inputs while maintaining speed efficiency.

\vspace{-2mm}
\section{Conclusions}\label{sec:6}
\vspace{-2mm}
% This paper presented a safety-critical control design approach for motion planning of an ASV operating in confined inland waterways. We leveraged the CBF framework to design a safety-critical RMPC scheme for ASVs. To address the safety issue on the inland waterways, we first formulated the safety functions required to avoid the border waterway and obstacles. The proposed safety functions were then adopted to define the corresponding discrete-time CBFs and embed them into the RMPC formulation. We finally demonstrated the performance of the proposed motion planning method and showed its merits compared to non-robust MPC, as well as RMPC without CBF constraints.
% As part of our future work, we intend to investigate the theoretical foundation of work. Future work will expand the proposed framework to address key challenges. This includes reducing computation time, extending it to multi-ASV systems, enabling cooperative navigation in congested waterways. Validating the approach in real-world experiments using physical ASVs will assess its performance under practical conditions. Incorporating dynamic obstacles and real-time environmental changes will enhance adaptability.
% By addressing these challenges, the proposed approach can advance toward large-scale deployment, paving the way for safer and more efficient autonomous systems on inland waterways.

This paper presents a safety-critical control design approach for motion planning of an ASV in confined inland waterways. We used the CBF framework to design a safety-critical RMPC scheme for ASVs. To address safety issues, we formulated safety functions to avoid the border waterway and obstacles. These functions defined discrete-time CBFs embedded in the RMPC formulation. We demonstrated the proposed method’s performance compared to non-robust MPC and RMPC without CBF constraints. 

Future work will investigate the theoretical guarantees, accelerating computation, and extending the scheme to cooperative multi-ASV navigation in dynamic environments to enable large-scale deployment. Furthermore, addressing dynamic obstacles and real-time changes will enhance adaptability and enable safe, large-scale ASV deployment.

\bibliography{IEEEabrv,references}

\end{document}